\pdfoutput=1

\documentclass[11pt]{article}

\usepackage[]{ACL2023}
\usepackage{amsfonts} 
\usepackage{graphicx}
\usepackage{hyperref}
\usepackage{times}
\usepackage{latexsym}
\usepackage{multirow}
\usepackage[T1]{fontenc}
\usepackage{booktabs}

\usepackage[utf8]{inputenc}

\usepackage{microtype}

\usepackage{inconsolata}

\title{MarsEclipse at SemEval-2023 Task 3: Multi-Lingual and Multi-Label Framing Detection with Contrastive Learning}

\author{Qisheng Liao \\
  MBZUAI \\
  \texttt{Qisheng.Liao@mbzuai.ac.ae} \\\And
  Meiting Lai \\
  Liaoning University \\
  \texttt{Lai.Meiting@outlook.com} \\\And 
   Preslav Nakov \\
  MBZUAI \\
  \texttt{Preslav.Nakov@mbzuai.ac.ae} \\}

\begin{document}
\maketitle
\begin{abstract}

This paper describes our system for SemEval-2023 Task 3 Subtask 2 on Framing Detection. We used a multi-label contrastive loss for fine-tuning large pre-trained language models in a multi-lingual setting, achieving very competitive results: our system was ranked first on the official test set and on the official shared task leaderboard for five of the six languages for which we had training data and for which we could perform fine-tuning. Here, we describe our experimental setup, as well as various ablation studies. The code of our system is available at \url{https://github.com/QishengL/SemEval2023}
\end{abstract}

\section{Introduction}
Framing involves highlighting certain elements of the perceived reality over others, which can promote problem definition, causal interpretation, moral evaluation, and/or treatment recommendation \cite{entman1993framing}. For example, events such as the outbreak of the COVID-19 pandemic can be discussed from different perspectives: health, economic, political, legal, quality of life, etc. Recently, machine learning has led to significant progress in framing detection. Previously, Na\"{i}ve Bayes, Support Vector Machines, and deep learning models based on recurrent neural networks (RNNs), long short-term memory networks (LSTM), and transformers have been used to analyze large amounts of text data and to capture complex patterns of framing \cite{morstatter2018identifying,liu2019detecting}.

Here, we describe our system for SemEval-2023 Task 3 \cite{semeval2023task3} Subtask 2 on Framing Detection, which is defined as a multi-label text classification task at the article level. The participants were required to build a system that can identify the frames used in an input article. Below, we describe our approach, which combines pre-trained language models (PLMs) and contrastive learning. Our contributions are as follows:

\begin{itemize}
\item We are the first to use contrastive learning in a multi-label framing detection setting.
\item We achieve very strong results: our system is ranked first for five out of the six languages for which training data was available, and it was third for English, which shows the effectiveness of contrastive learning for multi-lingual multi-label framing detection.
\item We perform ablation experiments to study the impact of different elements of our system.
\end{itemize}

\section{Related Work}
Contrastive learning has been used in computer vision for many years. It works as follows: a random example is picked as an anchor, and then positive and negative examples are selected with respect to that anchor. Next, the positive examples are pulled towards the anchor, while the negative ones are pushed away from it.

SimCLR is a popular contrastive learning model \cite{chen2020simple}, introduced in computer vision. It uses two views of the same image, generated using augmentations, and these are considered as positive examples. All other images are negative examples. This is a semi-supervised model, and it does not need labeled data for training. The result of this training is a good encoder with a strong capability for contrastive tasks. Then, the parameters are frozen, and the model is fine-tuned for a downstream task for which labeled data is available.
The loss function for contrastive learning is shown as equation \ref{eq:1} below. 
\begin{equation}\label{eq:1}
l_{i,j}=-\log\frac{\exp(sim(z_i,z_j)/\tau))}{\sum_{k=1}^{2n}\mathbb{L}[k\neq i]\exp(sim(z_i,z_k)/\tau))}
\end{equation}

Here, $i$ denotes an image, $j$ is a positive example for $i$, $k$ denotes all other images, $\tau$ is a temperature parameter, and $\mathbb{L}$ is an indicator function whose value is 1 if $k \neq i$ and it is 0 if $k=i$.

After SimCLR, a supervised contrastive learning model based on SimCLR was proposed by \citet{khosla2020supervised}. In supervised contrastive learning, each image is labeled. Images with the same labels are considered as positive examples, while such with different labels are taken as negative examples. The loss function is modified according to equation \ref{eq:2} below, where $P(i)$ is the set of all positive examples for $i$, and $A(i)$ is the set of all negative examples for $i$.
\begin{equation}\label{eq:2}
\sum_{i\in \mathbb{I}}\frac{-1}{\left| P(i) \right|}\sum_{j\in P(i)}\log\frac{\exp(sim(z_i,z_j)/\tau))}{\sum_{k\in A(i)}\exp(sim(z_i,z_k)/\tau))}
\end{equation}

Due to the success of contrastive learning in computer vision, many researchers have adapted it in natural language processing (NLP). SimCSE \cite{gao2021simcse} is a model derived from SimCLR. Instead of working with images, SimCSE operates on text inputs. The positive examples in SimCSE are generated by using dropout instead of other augmentations. The authors also proposed a method to weigh the negative examples. Some examples are considered as hard negatives and are pushed further apart from regular negative examples.

\begin{figure*}[h]
    \centering
    \includegraphics[width=0.8\textwidth]{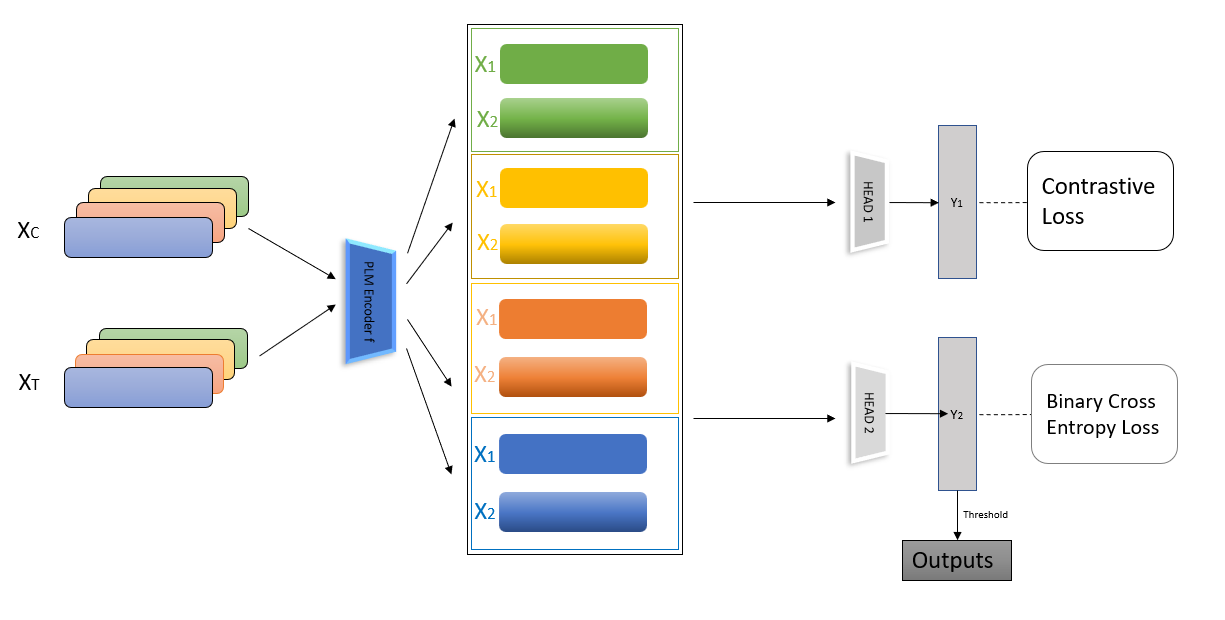}
    \caption{The architecture of our system. Different colors represent examples with different labels. The two inputs $X_C$ and $X_T$ are titles and bodies of articles. The encoder function $f$ is applied to both $X_C$ and $X_T$, generating two views of the representations for each article by means of dropout operations. Examples with the same color are considered positive, while such with different colors are considered negative.}
    \label{fig:con1}
\end{figure*}

\section{System Overview}

Our system uses XLM-RoBERTa as the backbone, and implements a multi-lingual and multi-label contrastive model on top of it. XLM-RoBERTa is well-suited for handling multi-lingual data. We trained a general model using data from various languages, with the aim of grouping examples with similar labels together, even if they are in different languages, and pushing apart examples with dissimilar labels. This strategy may allow the model to more easily identify a better decision boundary, as the number of clusters decreases. Moreover, by using contrastive loss, we expect increased robustness compared to models trained on just one language; it also helps fight class imbalance.

\begin{table}[]\centering
\begin{tabular}{llll}
\toprule
\bf Language & \bf Train & \bf Dev & \bf Test \\ 
\midrule
English  & 433   & 83  & 54   \\
French   & 158   & 53  & 50   \\
German   & 132   & 45  & 50   \\
Italian  & 227   & 76  & 61   \\
Polish   & 145   & 49  & 47   \\
Russian  & 143   & 48  & 72   \\
Spanish  & 0     & 0   & 30   \\
Georgian & 0     & 0   & 29   \\
Greek    & 0     & 0   & 64   \\ 
\bottomrule
\end{tabular}
\caption{Summary of the available data for subtask 2.}
\end{table}

\subsection{Input Representation}
We used two inputs: one from the title of the article and another one from the body of the article. We believe that the title of an article can often convey the main ideas discussed in the article, and thus can be a valuable source of information for frame detection. By incorporating both the title and the body of the article as inputs, our system can capture knowledge from multiple perspectives. In Section 5, we show that this helps the performance.

\subsection{Contrastive Loss}
We used a supervised architecture \cite{khosla2020supervised} to calculate the contrastive loss. While the task is a multi-label one and each example may have more than one label, we made some modifications to the conventional supervised contrastive loss equation and positive example definition. In particular, we considered two examples to be positive if they shared identical labels. If the examples have different labels, we calculated the number of distinct labels between them, denoted as $\Delta$. Then we passed $\Delta$ into a weight function $\mathbb{W}$ to determine the weight of a negative examples pair. $\mathbb{W}$ is a monotonically increasing function that assigns a higher weight to negative pairs with larger $\Delta$ values. The loss function is shown in equation \ref{eq:3}.
\begin{equation}\label{eq:3}
\resizebox{1\hsize}{!}{$\sum_{i\in \mathbb{I}}\frac{-1}{\left| P(i) \right|}\sum_{j\in P(i)}\log\frac{\exp(sim(z_i,z_j)/\tau))}{\sum_{k\in A(i)}\mathbb{W}(\Delta_{z_i,z_k})\exp(sim(z_i,z_k)/\tau))}$}
\end{equation}
\subsection{System Architecture}
Our system, depicted in Figure \ref{fig:con1}, takes two inputs for each example: its title $X_T$ and its body $X_C$. These are passed through a pre-trained language model encoder $f$, and the resulting outputs are concatenated to form $X_1$. We then generate another view of $X_1$ as $X_2$ using dropout, and we treat $X_1$ and $X_2$ as positive example pairs. These pairs are then fed into two separate heads. The first head produced outputs $Y_1$ that are used to compute the contrastive loss $L_{CL}$, while the second head produced outputs $Y_2$ that are used to compute the binary cross entropy loss $L_{CE}$. We then combine these losses into the final loss using a weight $\alpha$, as shown in equation \ref{eq:4}. Finally, we use a threshold on the outputs of $Y_2$ to obtain the final predictions.
\begin{equation}\label{eq:4}
L=\alpha L_{CE}+(1-\alpha)L_{CL}
\end{equation}

\subsection{Results Prediction \& Thresholds Selection}
As we mentioned above, the binary cross-entropy loss is used for optimization. Thanks to this loss function, we convert the multi-label classification problem into 14 binary classification problems as there are 14 different target frames to be predicted in subtask 2. Thus, for each input article, there are 14 logits corresponding to each of the 14 frames, and we predict that a given frame is to be assigned to the article if the value of the corresponding logit is higher than a certain threshold.

To find the optimal threshold for each language in the development set, we used a brute force algorithm that maximizes the F1 score. When testing on the six languages in the test set, we used the same thresholds determined by the brute force algorithm. For the three zero-shot languages, we set the threshold to 0.31, which was found to be the best threshold after averaging the best thresholds for all languages in the development set.

\subsection{Experimental Setup}
We used XLM-RoBERTa-Large \cite{liu2019roberta} as our encoder. During training, we used a batch size of 4, which was doubled to 8 for contrastive learning after generating two views of the inputs. We further used the Adam optimizer, with a learning rate of $1\times10^{-6}$. For simplicity, we used $\mathbb{W}(\Delta)=\Delta$ as our weight function . We performed the training on an NVIDIA A100 GPU.


\begin{table*}[]
\centering
\resizebox{2\columnwidth}{!}{%
\begin{tabular}{lccccccccccccccc}
\hline
 & \multicolumn{2}{c}{English}        & \multicolumn{2}{c}{French}         & \multicolumn{2}{c}{German}         & \multicolumn{2}{c}{Italian}        & \multicolumn{2}{c}{Polish}         & \multicolumn{2}{c}{Russian}        & \multicolumn{1}{l}{Spanish} & \multicolumn{1}{l}{Georgian} & \multicolumn{1}{l}{Greek} \\
                  & Dev            & Test              & Dev            & Test              & Dev            & Test              & Dev            & Test              & Dev            & Test              & Dev            & Test              & Test                        & Test                         & Test                      \\\hline
Baseline          & 0.605          & 0.349             & 0.380          & 0.328             & 0.506          & 0.487             & 0.430          & 0.485             & 0.592          & 0.593             & 0.215          & 0.229             & 0.120                       & 0.259                        & 0.345                     \\
Our System        & \textbf{0.753} & \textbf{0.562} \textit{(3)} & \textbf{0.611} & \textbf{0.552}\textit{(1)} & \textbf{0.636} & \textbf{0.711}\textit{(1)} & \textbf{0.618} & \textbf{0.617}\textit{(1)} & \textbf{0.690} & \textbf{0.673}\textit{(1)} & \textbf{0.543} & \textbf{0.449}\textit{(1)} & \textbf{0.477}\textit{(7)}              & \textbf{0.645}\textit{(2)}            & \textbf{0.497}\textit{(8)}                  \\ \hline

\end{tabular}%
}
\caption{The performance of our system compared to the baselines for all nine languages. The numbers in italic represent the ranking of our submissions on the official test leaderboard.}
\label{res}
\end{table*}

\section{Results}
Table \ref{res} shows the evaluation results for our best model, which we used to make predictions on the test data during the official testing phase for all nine different languages. Following the evaluation setup of subtask 2, we used F1 score as the official evaluation measure. We can see in the table that among the six languages for which we had training data and for which we could perform fine-tuning (English, French, German, Italian, Polish, and Russian), our system achieved 3rd place on the English leaderboard with an F1 score of 0.562. For the other five languages, our system was ranked 1st with F1 scores of 0.552, 0.711, 0.617, 0.673, and 0.449, respectively. Our system also achieved F1 scores of 0.477, 0.645, and 0.497 for the three zero-shot languages: Spanish, Georgian, and Greek, respectively.

\begin{table*}[]
\resizebox{2\columnwidth}{!}{%
\begin{tabular}{llllllllllllllll}
\hline
\multirow{2}{*}{} & \multicolumn{2}{c}{English}                                    & \multicolumn{2}{c}{French}                                     & \multicolumn{2}{c}{German}                                     & \multicolumn{2}{c}{Italian}                                    & \multicolumn{2}{c}{Polish}                                              & \multicolumn{2}{c}{Russian}                                    & Spanish                   & Georgian                           & Greek                     \\
                  & \multicolumn{1}{c}{Dev}            & \multicolumn{1}{c}{Test}  & \multicolumn{1}{c}{Dev}   & \multicolumn{1}{c}{Test}           & \multicolumn{1}{c}{Dev}   & \multicolumn{1}{c}{Test}           & \multicolumn{1}{c}{Dev}   & \multicolumn{1}{c}{Test}           & \multicolumn{1}{c}{Dev}            & \multicolumn{1}{c}{Test}           & \multicolumn{1}{c}{Dev}   & \multicolumn{1}{c}{Test}           & \multicolumn{1}{c}{Test}  & \multicolumn{1}{c}{Test}           & \multicolumn{1}{c}{Test}  \\
Our System        & \multicolumn{1}{c}{\textbf{0.753}} & \multicolumn{1}{c}{0.562} & \multicolumn{1}{c}{0.611} & \multicolumn{1}{c}{\textbf{0.552}} & \multicolumn{1}{c}{0.636} & \multicolumn{1}{c}{\textbf{0.711}} & \multicolumn{1}{c}{0.618} & \multicolumn{1}{c}{\textbf{0.617}} & \multicolumn{1}{c}{\textbf{0.690}} & \multicolumn{1}{c}{\textbf{0.673}} & \multicolumn{1}{c}{0.543} & \multicolumn{1}{c}{\textbf{0.449}} & \multicolumn{1}{c}{0.477} & \multicolumn{1}{c}{\textbf{0.645}} & \multicolumn{1}{c}{0.497} \\ \hline
\multicolumn{16}{c}{No Contrastive Learning}                                                                                                                                                                                                                                                                                                                                                                                                                                                                                  \\ \hline
No CL             & 0.751                              & \textbf{0.592}            & 0.628                     & 0.547                              & 0.662                     & 0.693                              & 0.605                     & \textbf{0.617}                     & 0.655                              & 0.656                              & 0.533                     & 0.412                              & 0.485                     & 0.540                              & 0.523                     \\ \hline
\multicolumn{16}{c}{Modify PLM}                                                                                                                                                                                                                                                                                                                                                                                                                                                                                               \\ \hline
Multilingual BERT & 0.697                              & 0.492                     & 0.598                     & 0.461                              & 0.645                     & 0.642                              & 0.580                     & 0.560                              & 0.661                              & 0.611                              & 0.547                     & 0.417                              & \textbf{0.540}            & 0.456                              & 0.481                     \\
XLM-RoBERTA-Base  & 0.733                              & 0.540                     & 0.622                     & 0.487                              & \textbf{0.670}            & 0.699                              & \textbf{0.622}            & 0.581                              & 0.667                              & 0.663                              & 0.515                     & 0.409                              & 0.486                     & 0.435                              & 0.526                     \\ \hline
\multicolumn{16}{c}{Use Whole Article as An Input}                                                                                                                                                                                                                                                                                                                                                                                                                                                                            \\ \hline
One Input         & 0.737                              & 0.572                     & \textbf{0.651}            & 0.485                              & 0.644                     & 0.638                              & 0.595                     & 0.561                              & 0.685                              & 0.595                              & 0.509                     & 0.400                              & 0.467                     & 0.525                              & \textbf{0.545}            \\ \hline
\multicolumn{16}{c}{Modify Negative Samples Weight}                                                                                                                                                                                                                                                                                                                                                                                                                                                                           \\ \hline
No Hard Examples  & 0.691                              & 0.566                     & 0.642                     & 0.540                              & 0.660                     & 0.696                              & 0.600                     & 0.607                              & 0.676                              & 0.667                              & \textbf{0.549}            & 0.421                              & 0.477                     &  \textbf{0.645}                     & 0.497                     \\ \hline
\end{tabular}%
}
\caption{The summary of our ablation study. Each row in the table is to be compared to the top row.}
\label{abalation}
\end{table*}

\section{Ablation Study}
In Table \ref{abalation}, we report our ablation results. All ablated test results are computed after the deadline of the shared task and are not reported on the official leaderboard.
We use XLM-RoBERTa-Large as an encoder, contrastive loss, two input features from each example, and hard negative examples for contrastive learning. The results of the ablation study showed that each component plays a role in our system.

\subsection{No Contrastive Learning}
As a first ablation, we trained a model without using contrastive loss. While it achieved the best score of 0.592 for English on the test data, the results for the other five fine-tuned languages were worse than for the full system. 
We observed that the number of training examples in English was 2 to 3 times greater than for the other languages. We concluded that when contrastive loss is not used, the model may be biased towards languages with more examples. In other words, contrastive loss helps the model to be more resilient to imbalanced datasets.

\subsection{Using a Smaller PLM}
We replaced the XLM-RoBERTa-Large encoder with Multilingual BERT \cite{devlin2018bert} base and with XLM-RoBERTa-Base. The results indicated that XLM-RoBERTa-Large was better for all six fine-tuned languages. Moreover, the results for XLM-RoBERTa-base are better than for Multilingual BERT base. We can conclude that a larger PLM yields better performance.

\subsection{Using an Entire Article as an Input}
We trained a model with an entire article as a single input as opposed to using our system, which takes the title and the body of an article as two inputs. We found that our system achieved better performance on testing for all six fine-tuned languages. We further observed that when using a single input, the model overfit severely on French in the development set.

\subsection{Changing the Negative Samples Weight}
We made a change to the weight function in our system, specifically setting $\mathbb{W}(\Delta)$ to be 1 for all negative samples, regardless of the number of differences between each negative pair. The results indicated that this modification led to worse performance for all languages in the testing phase. Interestingly, we also found that the testing results for the three zero-shot languages were exactly the same as those achieved by our proposed system.

\section{Conclusion and Future Work}
We have described our system for SemEval-2023 Task 3 Subtask 2, which used a combination of a PLM encoder and contrastive loss. On the official test set, our system was ranked 1st for five out of the six languages for which training data was available and thus we could perform fine-tuning, and we were ranked 3rd for English. We further explored alternative options for each component of our system and showed the individual contributions of each component.

We should note though that, while our system was the best overall on the six languages with training data, our results were not as good for the three zero-shot languages. 

In future work, we plan to explore ways to improve our model in a zero-shot learning scenario. We further plan to apply our model to other tasks, e.g., to subtask 3 of this SemEval-2023 task 3. It would be also interesting to experiment with adapters and to study their interaction with contrastive learning.

\bibliography{anthology,custom}
\bibliographystyle{acl_natbib}

\end{document}